\documentclass[conference]{IEEEtran}
\IEEEoverridecommandlockouts
\usepackage{cite}
\usepackage{amsmath,amssymb,amsfonts}
\usepackage{algorithm,algorithmic,amsmath}
\usepackage{graphicx}
\usepackage{textcomp}
\usepackage{xcolor}

\usepackage{lettrine}
\usepackage{wrapfig}
\usepackage{lipsum}
\usepackage{subfig}
\usepackage{booktabs}
\usepackage{multirow}
\usepackage{hhline}

\usepackage{varioref}
\usepackage{comment}
\usepackage{array}

\newcommand\tab[1][1cm]{\hspace*{#1}}
\usepackage{varioref}
\usepackage{hyperref}
\hypersetup{draft}
\usepackage{wrapfig}

\def\BibTeX{{\rm B\kern-.05em{\sc i\kern-.025em b}\kern-.08em
    T\kern-.1667em\lower.7ex\hbox{E}\kern-.125emX}}
\providecommand{\keywords}[1]
{
  \small	
  \textbf{\textit{Keywords---}}\textbf{\textit{#1}}
}

\makeatletter

\def\ps@IEEEtitlepagestyle{%
    \def\@oddfoot{\mycopyrightnotice}%
    \def\@evenfoot{}%
}
\def\mycopyrightnotice{%
    {\footnotesize  {\fontsize{8}{12}\fontfamily{ptm}\selectfont 978-1-5386-7097-2/18/\$31.00 \copyright 2018 IEEE\hfill }}
    \gdef\mycopyrightnotice{}
}



\makeatletter
\newcommand*\titleheader[1]{\gdef\@titleheader{#1}}
\AtBeginDocument{%
  \let\st@red@title\@title%
  \def\@title{%
    \bgroup\normalfont\large\centering\@titleheader\par\egroup
    \vskip1.5em\st@red@title}
}
\makeatother

\title{Deep Learning Approach for Building Detection in Satellite Multispectral Imagery}
\titleheader{ \raggedright {\fontsize{8}{12}\fontfamily{ptm}\selectfont 2018 International Conference on Intelligent Systems (IS) }}

\begin{document}

\author{
\IEEEauthorblockN{Geesara Prathap }
\IEEEauthorblockA{ \textit{Center for Technologies} \\ \textit{in Robotics and Mechatronics Components}
 \\ \textit{Innopolis University} \\ Innopolis, Russia \\
g.mudiyanselage@innopolis.ru}
\and
\IEEEauthorblockN{Ilya Afanasyev}
\IEEEauthorblockA{\textit{Institute of Robotics}
 \\ \textit{Innopolis University} \\ Innopolis, Russia \\
i.afanasyev@innopolis.ru}}
\maketitle

\begin{abstract}
Building detection from satellite multispectral imagery data is being a fundamental but a challenging problem mainly because it requires correct recovery of building footprints from  high-resolution images. In this work, we propose a deep learning approach for building detection by applying numerous enhancements throughout the process. Initial dataset is preprocessed by 2-sigma percentile normalization. Then data preparation includes ensemble modelling where 3 models were created while incorporating OpenStreetMap data. Binary Distance Transformation (BDT) is used for improving data labeling process and the U-Net (Convolutional Networks for Biomedical Image Segmentation) is modified by adding batch normalization wrappers.  Afterwards, it is explained how each component of our approach is correlated with the final detection accuracy. Finally, we compare our results with winning solutions of SpaceNet 2 competition for real satellite multispectral images of Vegas, Paris, Shanghai and Khartoum, demonstrating the importance of our solution for achieving higher building detection accuracy. \\
\end{abstract}


\section{Introduction}
%
%
%
%
The detection from satellite imagery becomes the actual problem for various urban applications: city planning, state cadastral inspection, infrastructure development, provision of municipal services, etc. Urban planners and state inspectors are responsible for strategic urban planning and long term development to reach proper sustainability and city growth. Subsequently, they are the ones who are responsible for detecting dwellings which have been built illegally. Although without having a proper observation of the area, this is a very challenging task. For instance, if there is a system which monitors the required area continuously over a period of time, it would help to determine whether it is an illegal project which should be terminated or not. 
%
%
%
%
%
%
%
%
%
%
%
%
%
%
%
%
Another example, radio, cellular and telecommunication companies need to make a proper decision about positioning their transmitter stations in order to achieve a wider coverage. Making an optimal configuration for broadcasting and communication transmitter stations highly depends on the profile of the terrain such as locations of existing buildings, lakes, rivers, etc. Thus, instead of dependence on mathematical modeling, a field survey is the best-adopted technique, but this is a very time-consuming approach. Almost all the scenarios explained above are to be dealt with either spatial or spatial-temporal processes that evolves in time.
  Remote Sensing (RS) is composed of a set of techniques and methods to explore the Earth’s surface at a long distance, where the object detection is the most prominent task as well as the most difficult to solve. According to~\cite{Abburu2015SatelliteIC}, object detection in RS broadly classifies into three categories: automatic, manual and hybrid. In recent times, significant attention is received for automatic detection because of various reasons such as the revolution of the deep neural network approaches for object detection, dramatic increase of computing power with respect to the cost, etc. DaoYu Lin ~\cite{lin2016deep} has proposed one of the deep unsupervised representation learning techniques which is one of the successful works in the RS context. But in this work also, some improvements can be done to increase the final accuracy. Another example is presented in ~\cite{ishii2016detection}, where training a CNN from scratch to classify multispectral image patches taken by satellites as whether or not they belong to a class of buildings was realized. This additional information will help to narrow down the solution space in an optimal way. More information can be found in~\cite{dornaika2016building}, ~\cite{huang2014multi}. Input feature descriptor should be decided by the model itself without making any statistical hypothesis beforehand based on the dataset only.
The main motivation of this research is to investigate possible approaches to automatically detect a mapping between input and desired output. The deep neural network should decide itself what are the most appropriate feature descriptors to map between input and output that maximizes the final accuracy of the model. 
%
%
%
%
%
%
%
%

Utmost important step of satellite imagery when detecting objects is the mapping of actual earth surface locations into image pixel values in a meaningful way. Initially, there are a few meta data available for processing such as Ground Truth Distance (GTD) and bands which belong to images. 
%
%
%
%
%
%
%
%

The success behind the winning solution~\cite{krizhevsky2012imagenet} for the Larger Scale Visual Recognition Challenge (ILSVRC)~\cite{everingham2010pascal} is to apply CNNs (Convolutional Neural Networks) that have resulted in various solutions proposed for object detection. One satisfactory solutions is described in the article ~\cite{ostankovich2018}, where some computer vision techniques for building detection are integrated for generating region proposals, which are classified by retrained GoogLeNet CNN with the following building legality validation with a state cadastral map.
The paper~\cite{girshick2014rich} presents another good example with R-CNN, although main drawbacks of similar technique have been identified in~\cite{krizhevsky2012imagenet} with proposals how to mitigate them in a proper way. This proposed solution has two key insights: (1) CNN is employed for localizing and segmenting objects, and (2) applies supervised pre-training as a supplementary task. 
%
%
%
%
%
%
%
%
%
%
%
%

Performing initial dataset normalization and changing the image resolution may help for increasing detection accuracy. Typically, resolution of satellite imagery is low because of complex atmospheric conditions, which influence on spatial, spectral, radiometric and temporal characteristics. In\cite{genitha2010super}, Hopfield neural network is used to increase the spatial resolution of images. On the other hand,  interpolation method\cite{qifang2017super} is employed for reconstruction of super-resolution images. Since dataset normalization solely depends on the given dataset, we proposed a technique for selected dataset in a proper way.   

Our proposed solution has been framed, after considering all those facts.
%
%
%
%

\textbf{Our contributions}
\begin{enumerate}
\item Proposing an ensemble modeling for building footprints detection by using multi-spectral satellite images and open street map data. 
 \begin{itemize} 
        \item Selecting suitable bands (Coastal, Yellow, Red, Red
Edge and Near-IR1) for building footprints. These were selected by empirical testing. 
        \item To reduce the computational power resizing images into 256x256 resolution. On the contrary, to retrieve the loss information after resizing, the second
model where each image with an original resolution (650x650) slice into 9 images is employed.
        \item To overcome suspected false positive during the labelling and improve a sharpness of building footprints, model 3 is introduced by incorporating 2
layers (building and roads) form the open street map data.
     \end{itemize}

\item 2-Sigma percentile normalization for the initial dataset normalization.
\item BDT~\cite{neiva2016binary} to improve the building footprint labeling accuracy.
\item Modifications for the U-Net~\cite{ronneberger2015u}:
 \begin{itemize} 
        \item Introducing batch normalization wrapper around the activation function of each layer.
        \item Stochastic Gradient Descent (SGD) algorithm which is used as the optimizer for original U-Net is changed into ADAM~\cite{kingma2014adam} optimizer to increase the final accuracy.
     \end{itemize}
\end{enumerate}

\section{Methodology} \label{sec:DatasetAnalysis}

\subsection{Dataset Selection} \label{sec:DatasetSelection}

When selecting a dataset, these criteria should be considered: number of images, feasibility of quick and accurate labeling of dataset, information available on bands (e.g. RGB or multi spectral information), image resolution and information about correspondence between ground truth and pixel location. After considering all those facts SpaceNet\footnote{SpaceNet dataset: \url{https://github.com/SpaceNetChallenge}\label{SpaceNet}} dataset has been selected. It comprises commercial satellite imagery and labeled training data. It provides dataset  of four cities: Vegas, Paris, Shanghai and Khartoum in the following format: \\
%
%
%
%
\tab\textbf{geojson}- GeoJson labels of buildings for each tile\\
\tab\textbf{MUL}- 8-Band Multi-Spectral images \\
\tab\textbf{MUL-PanSharpen}- 8-Band Multi-Spectral images\\
\tab\textbf{PAN}- Panchromatic images\\
\tab\textbf{RGB-PanSharpen}- RGB images\\
\tab\textbf{summaryData}- pixel based labels for the building\\
Pan sharpen images consist of merging high-resolution panchromatic and low resolution multi spectral images.   To create a single high-resolution color image, 8-band images have not been pan-sharpened. Building footprint corresponding to each image can be found in $summaryData$ in this format: $ImageId,BuildingId,PolygonWKT\_Pix$ where $PolygonWKT\_Pix$ stands for defining building footprint location in Well-Known Text (WTK) format.
%
%

\subsection{Data Preprocessing} \label{sec:DataPreprocessing}

The whole dataset is divided into two separate sets: training(80\%) and testing(20\%). Again training dataset is divided into two separate sets: training (70\%) and validating (30\%). 
%
\begin{figure}[!ht]
\begin{center}
\includegraphics[width=\linewidth, height=8cm]{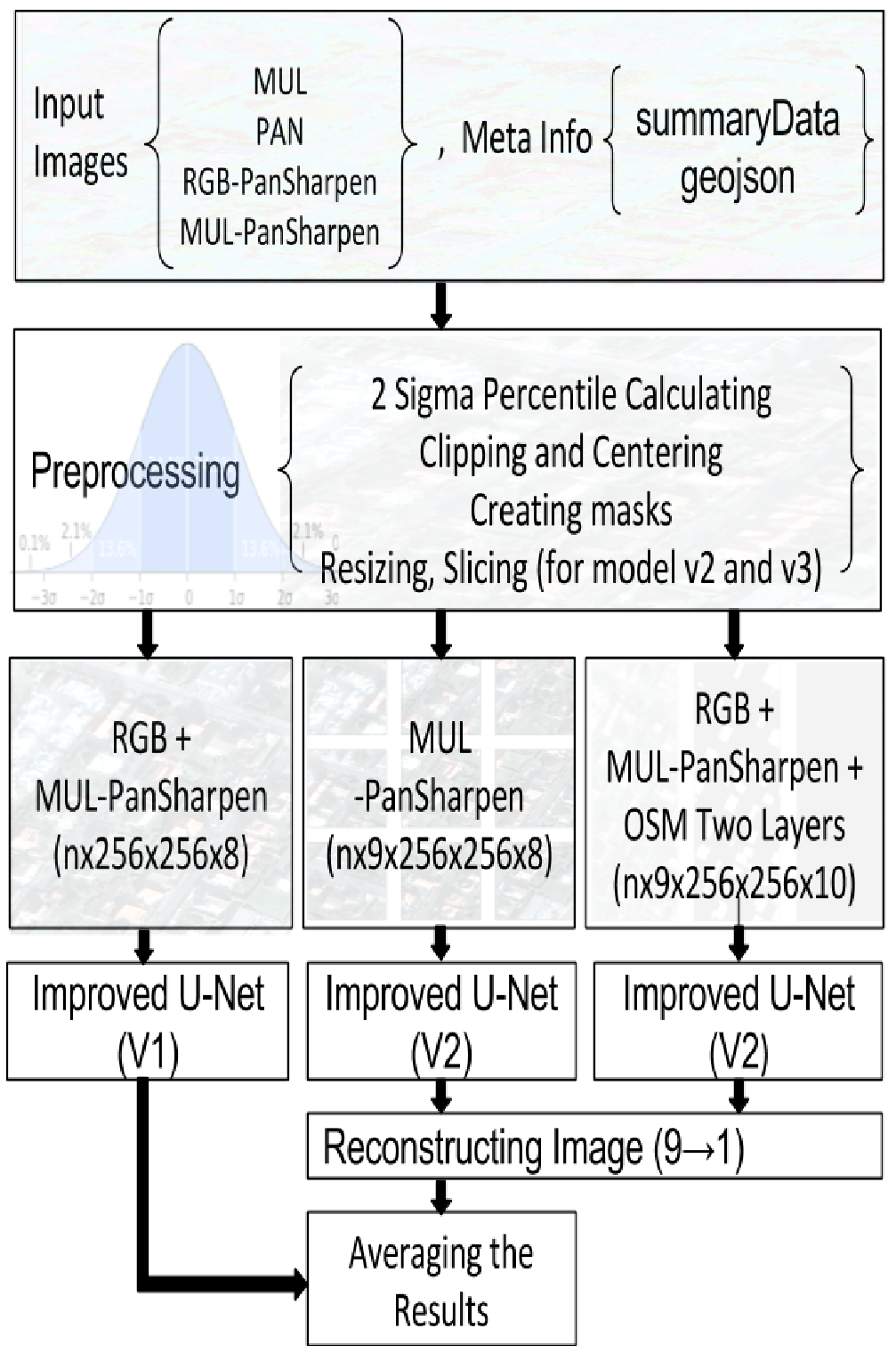}
\caption{\label{f:Processing_in_brief} {\fontsize{8}{12}\fontfamily{ptm}\selectfont The high-level architecture for the proposed building detection methodology, which contains 4 stages: preprocessing, image preparation, model training and building footprint prediction.}}
\end{center}
\end{figure}
RGB-PanSharpen images consist of three bands which are Red, Green and Blue. Multi-PanSharpen images consist of 8 bands~\cite{DGWorldView3DS2014}: Coastal (400 - 450 nm), Blue (450 - 510 nm), Green (510 - 580 nm), Yellow (585 - 625 nm), Red (630 - 690 nm), Red Edge (705 - 745 nm), Near-IR1 (770 - 895 nm), Near-IR2 (860 - 1040 nm). Then, basic statistical properties (mean, standard deviation, max and min) corresponding to each channel (bands) are calculated for all training and validating images considering as a single dataset. After making a hypothesis such that each channel statistics belongs to a form of a normal distribution, $-2\sigma$ and $+2\sigma$ where $\sigma$ is the standard deviation values which corresponds to each channel is calculated. $-2\sigma$ and $2\sigma$ are the upper and lower bound on each channel for the whole dataset. However, -2$\sigma$ is the 2.28th percentile value and +2$\sigma$ is the 97.72nd percentile value. Next step is to clip images which exceed upper limit into  +2$\sigma$ percentile value and fall behind the -2$\sigma$ into -2$\sigma$ percentile value. In this way noise of each channel can be reduced. This process is shown in Fig.~\ref{f:Processing_in_brief}. Each channel is normalized based on the min-max normalization technique as follows: 

\begin{equation}
image\_channel_i = \frac{images\_channel_i - min(image\_channel_i) }{ max(image\_channel_i)-min(images\_channel_i)}
\end{equation}

\begin{figure}
\centering
\subfloat[][]{\includegraphics[width=2.7cm, height=2.7cm]{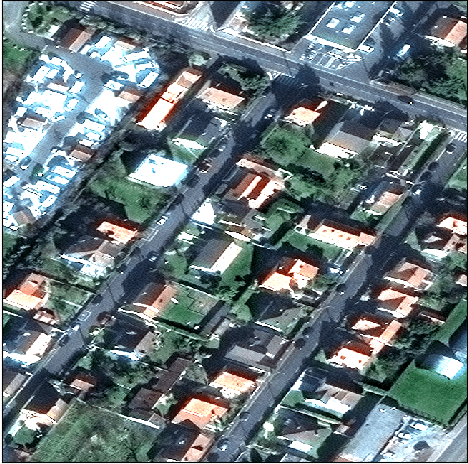}\label{f:v2input}}
\vspace{1pt}
\subfloat[][]{\includegraphics[width=2.7cm, height=2.7cm]{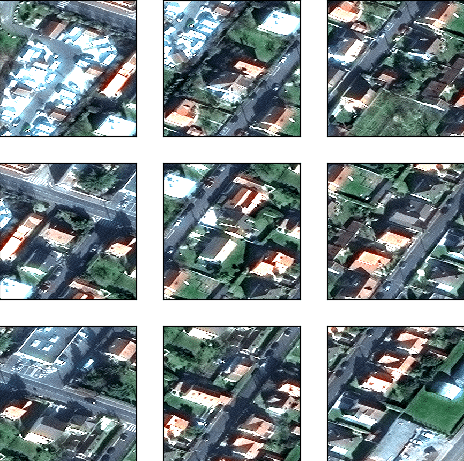}\label{f:v2resized}}
\caption{\protect\subref{f:v2input}  {\fontsize{8}{12}\fontfamily{ptm}\selectfont  Normalized input image with 256x256 resolution (only RGB bands are shown), 
  \protect\subref{f:v2resized} Sliced images (only RGB bands are shown).}}
\end{figure}

\subsection{Model Building} \label{sec:ModelBuilding}

One of the investigations (U-Net~\cite{ronneberger2015u}) which won the ISBI tracking challenge 2015~\cite{ISBIChallenge} is used for segmentation of neuronal structures in electron microscopic stacks. Although this work was applied in different context, building footprints can be imagined as neuronal structures. Thus, U-Net can be directly mapped for segmenting out building footprints. Initially there are 8 or 10 feature maps, which correspond to each input image with 256x256 resolution, enter to the network which is shown in Fig.~\ref{f:Processing_in_brief} next to the prepossessing stage. 

The main idea of U-Net compared to ~\cite{long2015fully} is replacing pooling operators between successive layers by upsampling operators. Hence, these layers increase the output resolution. This technique helps to localize desired objects with a high accuracy because when doing the upsampling, total feature set or image depth is increased that will help to create a proper mapping between input and the output. In addition, this network only uses the valid part of each convolution with no fully connected layers throughout the network. 

Most deep neural networks including U-Net are facing the following common problem:  distributions of the output of each hidden layer always depend on the result of previous layers. Thus, model parameters are changed during the learning process. Each layer needs to adapt to those changes during the training process. To address this problem, Sergey Ioffe and Christian Szegedy suggested a simple and effective solution in~\cite{ioffe2015batch}, which is called the batch normalization. The basic idea proposed by this solution is to wrap batch normalization layer around input to the activation function. As an example, $\sigma(Wx + b)$ is the activation function such as sigmoid or Rectified Linear Unit (ReLU). The input should be wrapped with batch normalization (BN) in this manner: $\sigma(BN(Wx + b))$. Hence, U-Net is modified by applying batch normalization. 
%
%
%
%

Model accuracy is calculated in the training phase by using Jaccard Index ($J(A,B)$) between area $A$ and $B$ which can be denoted as: 
\begin{equation}
J(A,B) = \frac{|A \cap B|}{ |A| + |B| - |A \cap B|}
\end{equation} where $A$ is the ground truth and $B$ is the predicted area by the model.  

The optimizer also completely affects the accuracy of the final model. Original U-Net uses the stochastic gradient descent algorithm (SGD). However, in this implementation ADAM ~\cite{kingma2014adam} optimizer is being used instead. The main reason for selecting ADAM over SGD is to increase the model accuracy.  According to ~\cite{kingma2014adam}, it is demonstrated that ADAM outperforms problems for noisy data and non-stationary objectives. In addition, it uses moving average of the momentum which enables the use of large effective step size. The algorithm will take care of the convergence even it takes some considerable amount of time because the prime goal of this research is to improve the building detection accuracy.     
%
%
%
%

An ensemble modelling of building footprints detection contains three models which are denoted by v1, v2 and v3. Following subsections will explain more details on each model. 

\subsubsection{First Model (v1)}

\begin{figure}[!ht]
\begin{center}
\includegraphics[width=6.6cm, height=2.9cm]{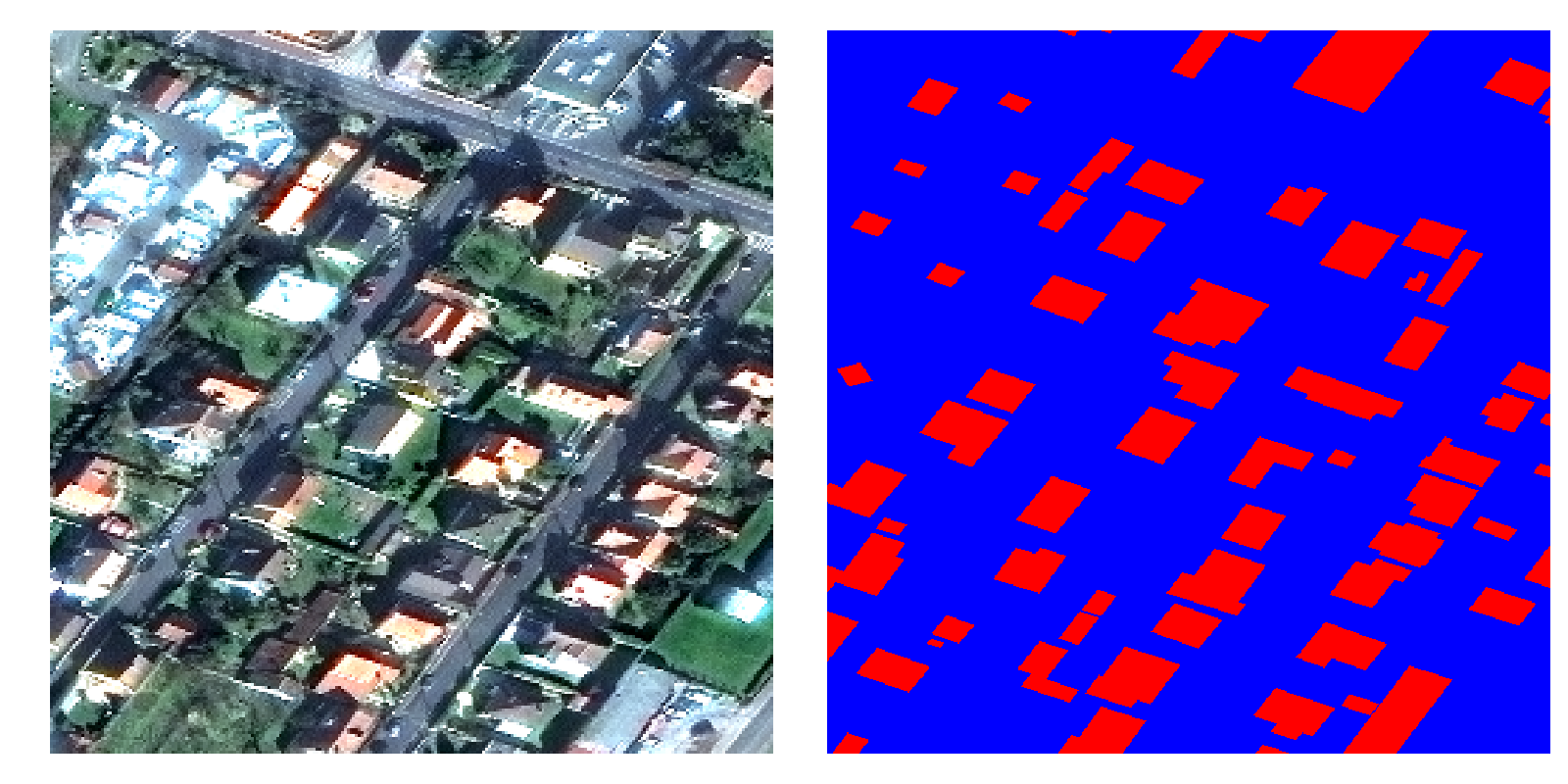}
\caption{\label{f:apply_mask} {\fontsize{8}{12}\fontfamily{ptm}\selectfont After locating the building footprints by using the given ground truth. However, the main problem with this technique is that it may badly affect the small building as seen in this figure.}}
\end{center}
\end{figure}

\begin{figure}[!ht]
\begin{center}
\includegraphics[width=6.6cm, height=2.9cm]{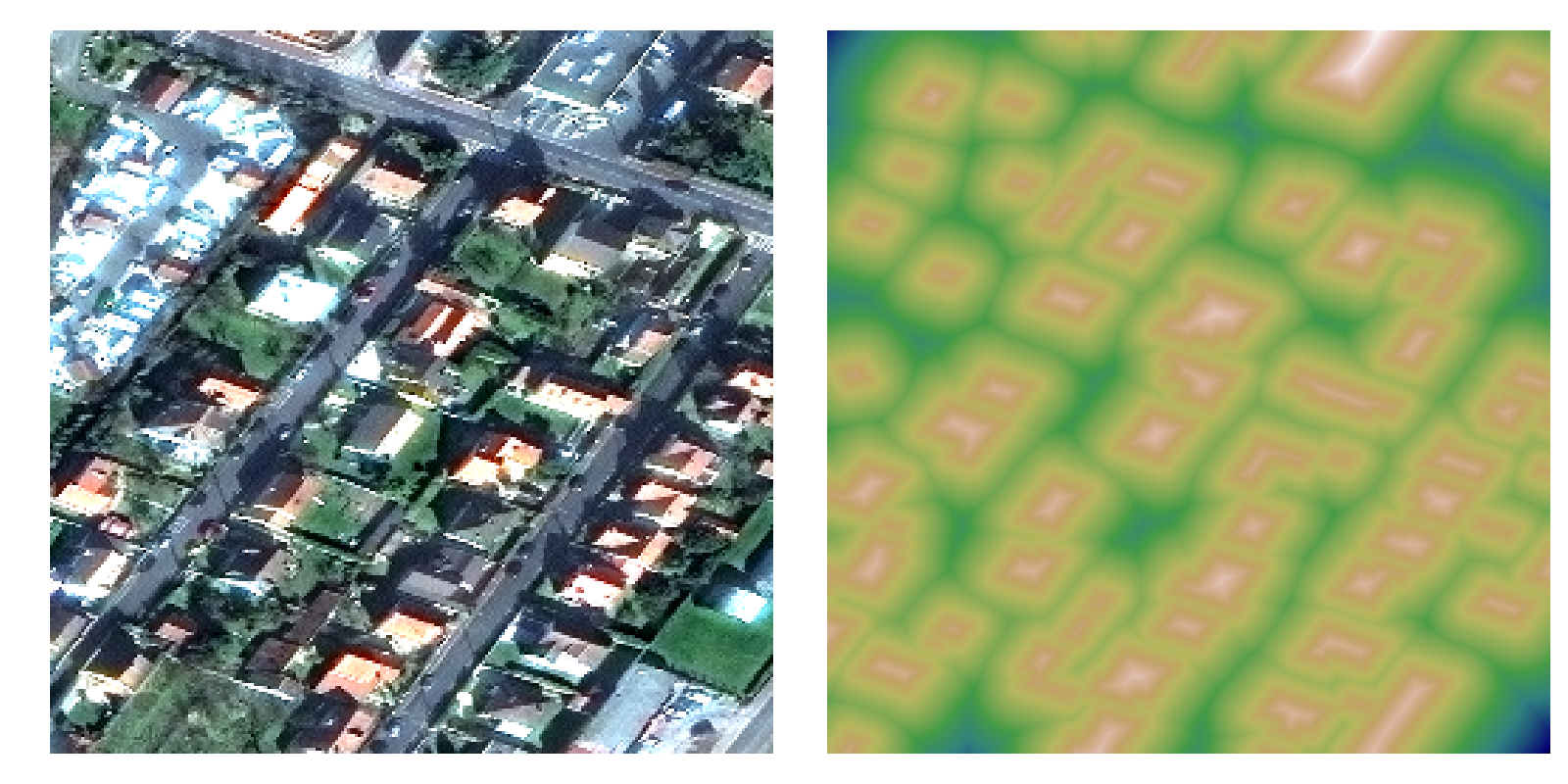}
\caption{\label{f:apply_distance_transformation} {\fontsize{8}{12}\fontfamily{ptm}\selectfont After applying binary distance transformation suggested by ~\cite{neiva2016binary}. This process is important when the building footprints' sizes are quite small.}}
%
%
\end{center}
\end{figure}

Input of the v1 contains 8 bands: 3 bands form RGB-PanSharpen and 5 bands from Multi-PanSharpen images which include Coastal, Yellow, Red, Red Edge and Near-IR1. Creating label corresponding to each input image should be done with a 650x650 resolution where building footprint of each image is provided. Basic steps which involve data labeling images are
\begin{enumerate}
\item Reading building locations which are corresponding to each training images
\item Initializing 650x650 matrix with zeros
\item Drawing polygons which are extracted from the first step as shown in Fig.~\ref{f:apply_mask}
\item Applying BDT: a positive distance is calculated if pixels inside buildings, otherwise a negative distance will be instead. Result of this transformation is shown in Fig.~\ref{f:apply_distance_transformation}.
\end{enumerate}
After creating labeled images, it is needed to be resized into desired resolution (256x256x1).

\subsubsection{Second Model (v2)}

Resizing image resolution leads to destroy its useful information that is inevitable. To overcome this issue, the second model is being employed. Basic steps for creating input images and labels are the same as v1. However for v2 input image (e.g Fig.~\ref{f:v2input}) is sliced into 9 images as shown in Fig.~\ref{f:v2resized} with 256x256 resolution. Multi-PanSharpen images are used in v2 and creating masks is equivalent to the process is done in model v1 corresponding to each sliced image.

\subsubsection{Third Model (v3)}

OpenStreetMap (OSM)\footnote{OpenStreetMap dataset: \url{https://mapzen.com/}\label{OSM}}
data is usually used for semantic labelling in RS for various purpose such as a ground truth, automatic labelling, etc. In~\cite{audebert2017joint}, OSM is used for automatic object extraction or to build better semantic maps with the help of deep learning techniques. v3 applies the same concept. OSM shape files consist of various type of layers including buildings, lands industrial zones, roads and water resources. Buildings and roads are extracted corresponding to provided PAN images with the intention of sharpening input images. To extract required polygons from shape files, it is necessary to transform the image coordinate system into latitudes and longitudes where buildings and roads can be extracted from the shape files. 

\begin{figure}[!ht]
\begin{center}
\includegraphics[width=9cm, height=2.4cm]{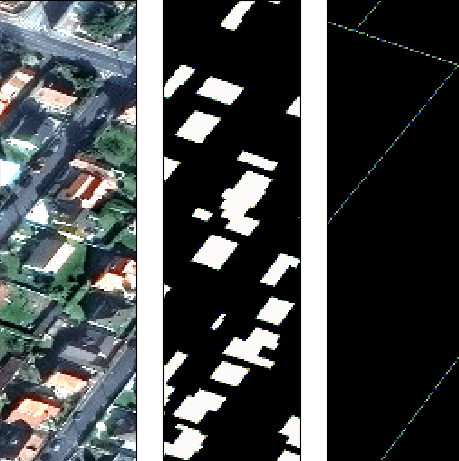}
\caption{\label{f:osmlayers} {\fontsize{8}{12}\fontfamily{ptm}Normalized input image and extracted OSM layers (buildings and roads) corresponding to geolocation of the input image.} }
\end{center}
\end{figure}

Extracted shapes should be transformed back to the image coordinate system where required objects can be localized within the image. How OSM can be utilized to extract required layers is shown in Fig. ~\ref{f:osmlayers}. Extracted layers are added to each image and sliced into 9 images with 256x256 resolution which is the same process employed in the v2 building. In addition to v2, v3 comprises 2 additional layers extracted from OSM.

Once images are constructed for each model (v1, v2 and v3), normalization is done by calculating mean image for whole training dataset corresponding to each model and subtracting each image by corresponding mean image. This process can be seen as centering data around zero. This also allows coherent and aligned handling of noises in an optimal manner.
%
%

\section{Model Evaluation} \label{sec:ModelEvaluating}

As explained in the section \ref{sec:ModelBuilding}, there is an image set which dedicated to model validation. Initial model training, 300 epoch along with 64 batch size is used in each model. End of each epoch, an accuracy of the model is calculated using validating dataset. To measure the proximity between labeled building footprints and predicting footprints of each model, Intersection of Union (IoU)~\cite{choi2010survey} similarity measure can be defined as: 

\begin{equation} 
\textbf{IoU} = \frac{ Area(A \cap B) }{ Area(A \cup B)}
\end{equation} where A and B are corresponding to labelled and predicted building polygon locations. Value of IoU is varied between 0 and 1. If the result is close to one, prediction accuracy is high. Otherwise it has low accuracy. To identify whether a detected polygon is a true positive or false positive, IoU is used. If measured IoU is 0.5 or higher, it is considered as a true positive, otherwise, it is considered as a false positive. 

Once clearly separating out whether it is false positive or true positive, model accuracy can be calculated using \textbf{F score} which is inspired by ~\cite{russakovsky2015imagenet}. F score can be seen as a harmonic mean of precision and recall because it comprises the accuracy of precision measure and the completeness of the recall measurement. F score can be defined as:
\begin{equation} \textbf{F1 score} = 2*\frac{ precision*recall }{ precession+recall} = 2*\frac{ \frac{tp}{M}* \frac{tp}{N}}{ \frac{tp}{M}+\frac{tp}{N}} = \frac{2*tp}{M+N}\end{equation} where N is the number of polygon labels for building footprints, which is considered as ground truth, and M denotes the proposed polygons by each model. Here $tp$ denotes the number of true positives out of the proposed polygons (M). Value of F score is varied between 0 and 1, where greater value means better the result. 
%
%

The final result is calculated by averaging the results of model v1, v2 and v3. In the training phase, model parameters were saved corresponding to the highest F score for each model. Detected masks need to be constructed by combining 9 sub-images corresponding to each image. This process is applied only for model v2 and v3. The result of v1, v2 and v3 are summed and divided by the number of images which applies to each pixel location. The result of this is the final predicted building footprints. Since original dataset is provided with 650x650 resolution, predicted result resized into original resolution. There are some detection area that can be intercepted with a few other polygons. Those polygons are cascaded into one. The result of this process is a sequence of polygons where the area is greater than given minimum threshold ($minArea$). If the size of polygon area is less than $minArea$, those are neglected. Finally, the result is stored in this format: $ImageId,BuildingId,PolygonWKT\_Pix$.
%
%

\section{Evaluation Results and Analysis}

Final evaluation summary of each city is shown in the table ~\ref{t:evaluationmatrix}. It can be clearly seen that final results are varied from city to city. The main reason can be that our proposed solution does not handle very small building footprints very accurately. As an example when comparing Vegas or Paris and Shanghai or Khartoum most buildings which are located in Vegas and Paris are quite bigger than Shanghai and Khartoum. That is why there is approximately 0.2 difference in F score between these pairs of cities.  When comparing with the winning solutions of SpaceNet Round2 our solution is able to get maximum F score on each city (see, the table ~\ref{t:finalresult}). 
%
%

\begin{table}[h!]
\centering
\captionsetup{font=scriptsize}
\caption{ {\fontsize{8}{12}\fontfamily{ptm}\selectfont \textsc{THE BEST RESULT OF OUR PROPOSED SOLUTION FOR EACH CITY CORRESPONDING TO THE MINIMUM POLYGON AREA}}}
\label{t:evaluationmatrix}
\begin{tabular}{|l|l|l|l|l|l|l|}
\hline
City      & Precision & Recall & F-score & minArea\\ \hline
Vegas   & 0.9300       & 0.8420     & 0.8838 &  120  \\ \hline
Paris   & 0.8277        & 0.7031    & 0.7603  & 180 \\ \hline
Shanghai & 0.6832        & 0.5022     & 0.5789   & 180 \\ \hline
Khartoum &0.7031 & 0.5303 & 0.6045 & 180\\ \hline
\end{tabular}
\end{table}

\begin{table}[h!]
\centering
\captionsetup{font=scriptsize}
\caption{{\fontsize{8}{12}\fontfamily{ptm}\selectfont \textsc{COMPARISON BETWEEN OUR PROPOSED SOLUTION AND PROVISIONAL SCORES FOR WINNERS OF SPACENET ROUND2}}}
\label{t:finalresult}
\begin{tabular}{|l|l|l|l|l|}
\hline
Competitor   & Las Vages & Paris    & Shanghai & Khartoum \\ \hline
XD\_XD       & 0.885     & 0.745    & 0.597    & 0.544    \\ \hline
wleite       & 0.829     & 0.679    & 0.581    & 0.483    \\ \hline
nofto        & 0.787     & 0.584    & 0.520    & 0.424    \\ \hline
Our Solution & \textbf{0.883}  & \textbf{0.760} & \textbf{0.604} & \textbf{0.584} \\ \hline
\end{tabular}
\end{table}

\begin{figure}
  \subfloat[]]{%
  \begin{minipage}{\linewidth}
  \includegraphics[width=4.38cm, height=4.5cm]{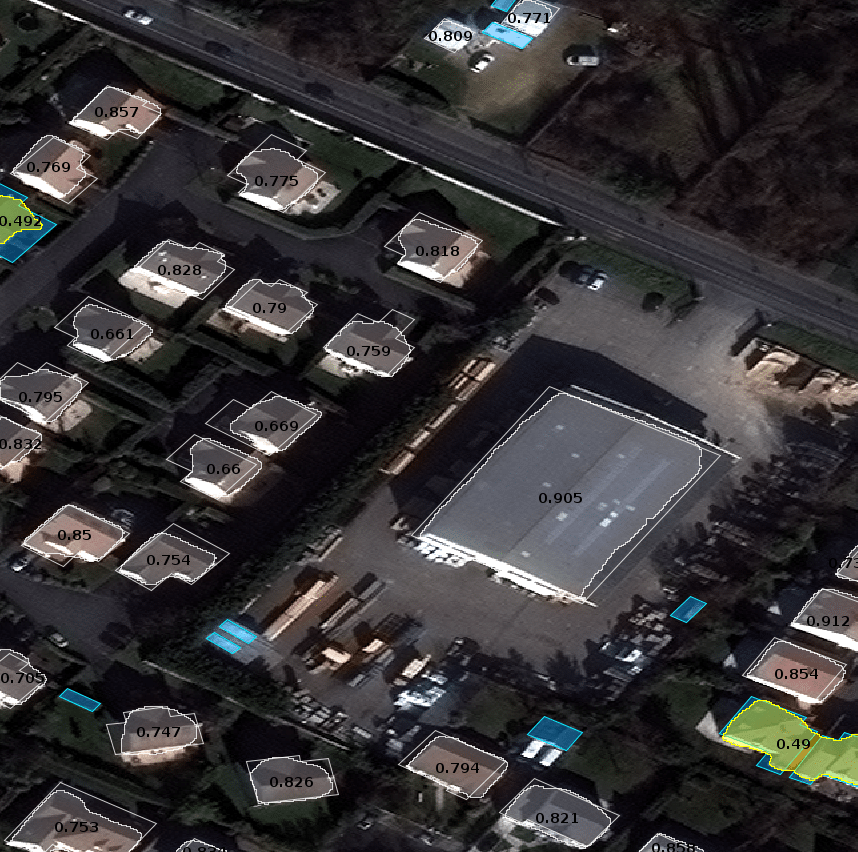}\hfill
  \includegraphics[width=4.38cm, height=4.5cm]{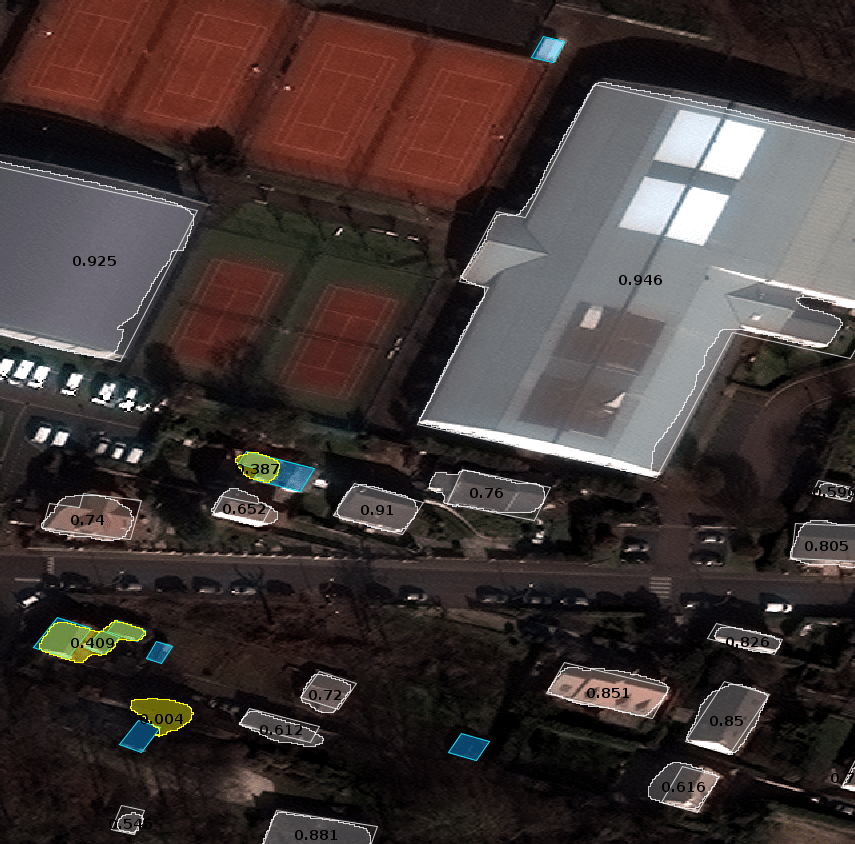}%
  \end{minipage}%
  }\par
  \vspace{3pt}
  \subfloat[]]{%
  \begin{minipage}{\linewidth}
  \includegraphics[width=4.38cm, height=4.5cm]{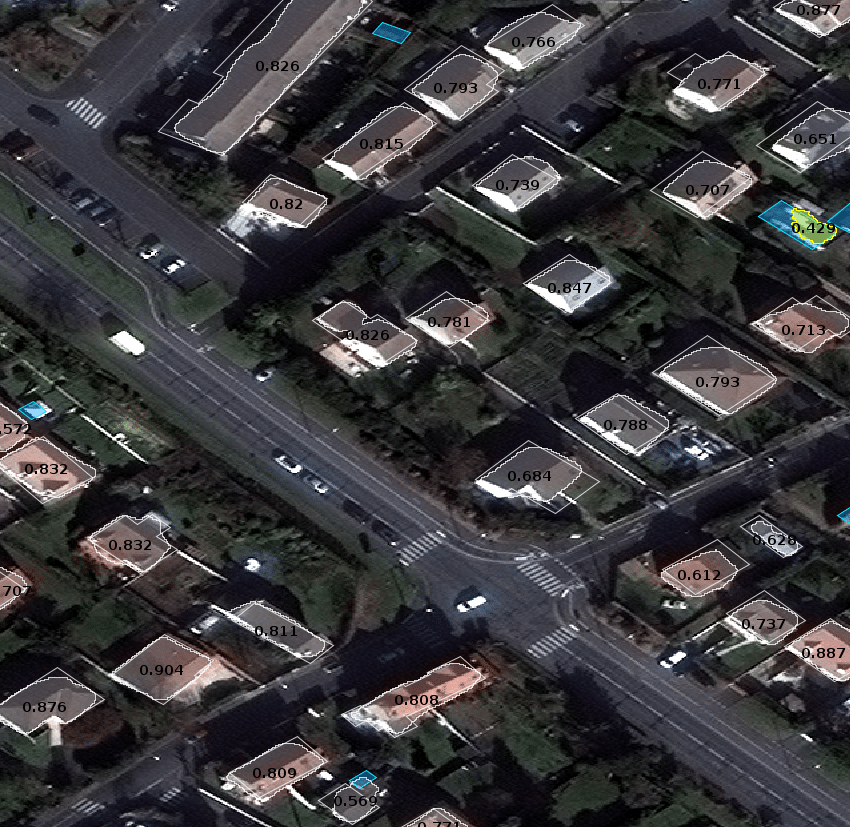}\hfill
  \includegraphics[width=4.38cm, height=4.5cm]{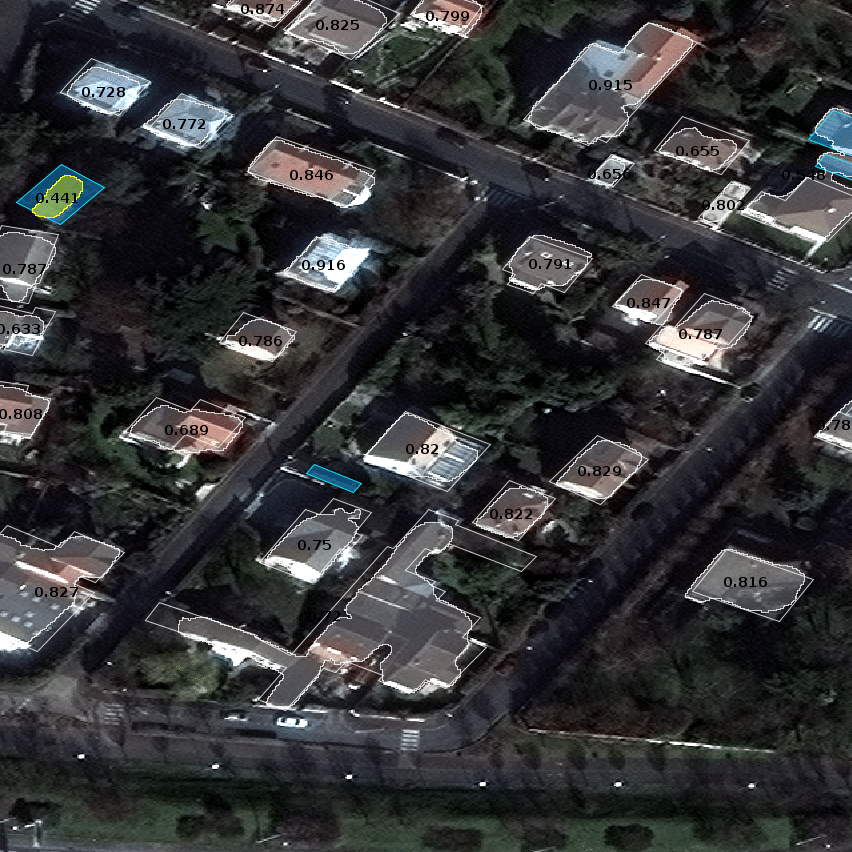}%
  \end{minipage}%
  }\par
  \caption{\fontsize{8}{12}{\fontfamily{ptm}\selectfont Final building footprints detection result with F score by using our proposed solution.}}\label{fig:finalresuts}
\end{figure}

\section{\textbf{Conclusions}}\label{sec:Conclude}

In this letter, a new framework has been proposed to detect building footprints for a given dataset. Proposed solution is outperformed when the size of the building is quite big. But for very small building footprints, model does not succeed to detect correctly that should be improved. Mainly there are two approaches that can be considered in order to overcome those issues. The first one is to select the training dataset which has more variance then build the model more dipper. The second approach would be incorporating unsupervised deep learning approaches such as Generative Adversarial Networks it order to train the model. 
%
%
%
%
%
%
%
%


%

\section*{Acknowledgment}
The authors would like to thank Kohei Ozaki\footnote{Winning Solution for the SpaceNet Challenge: \url{https://i.ho.lc}\label{Kohei}} who shared the winning solution of the SpaceNet Challenge Round 2, whose initial idea about building detection inspired to our methodology. The authors are also grateful to Innopolis University for the support.

\ifCLASSOPTIONcaptionsoff
  \newpage
\fi



\bibliographystyle{IEEEtran}
\bibliography{IEEEexample}

\begin{thebibliography}{10}
\providecommand{\url}[1]{#1}
\csname url@samestyle\endcsname
\providecommand{\newblock}{\relax}
\providecommand{\bibinfo}[2]{#2}
\providecommand{\BIBentrySTDinterwordspacing}{\spaceskip=0pt\relax}
\providecommand{\BIBentryALTinterwordstretchfactor}{4}
\providecommand{\BIBentryALTinterwordspacing}{\spaceskip=\fontdimen2\font plus
\BIBentryALTinterwordstretchfactor\fontdimen3\font minus
  \fontdimen4\font\relax}
\providecommand{\BIBforeignlanguage}[2]{{%
\expandafter\ifx\csname l@#1\endcsname\relax
\typeout{** WARNING: IEEEtran.bst: No hyphenation pattern has been}%
\typeout{** loaded for the language `#1'. Using the pattern for}%
\typeout{** the default language instead.}%
\else
\language=\csname l@#1\endcsname
\fi
#2}}
\providecommand{\BIBdecl}{\relax}
\BIBdecl

\bibitem{Abburu2015SatelliteIC}
S.~Abburu and S.~B. Golla, ``Satellite image classification methods and
  techniques: A review,'' 2015.

\bibitem{lin2016deep}
D.~Lin, ``Deep unsupervised representation learning for remote sensing
  images,'' \emph{arXiv preprint arXiv:1612.08879}, 2016.

\bibitem{ishii2016detection}
T.~Ishii, E.~Simo-Serra, S.~Iizuka, Y.~Mochizuki, A.~Sugimoto, H.~Ishikawa, and
  R.~Nakamura, ``Detection by classification of buildings in multispectral
  satellite imagery,'' in \emph{Pattern Recognition (ICPR), 2016 23rd
  International Conference on}.\hskip 1em plus 0.5em minus 0.4em\relax IEEE,
  2016, pp. 3344--3349.

\bibitem{dornaika2016building}
F.~Dornaika, A.~Moujahid, Y.~El~Merabet, and Y.~Ruichek, ``Building detection
  from orthophotos using a machine learning approach: An empirical study on
  image segmentation and descriptors,'' \emph{Expert Systems with
  Applications}, vol.~58, pp. 130--142, 2016.

\bibitem{huang2014multi}
X.~Huang, Q.~Lu, and L.~Zhang, ``A multi-index learning approach for
  classification of high-resolution remotely sensed images over urban areas,''
  \emph{ISPRS Journal of Photogrammetry and Remote Sensing}, vol.~90, pp.
  36--48, 2014.

\bibitem{krizhevsky2012imagenet}
A.~Krizhevsky, I.~Sutskever, and G.~E. Hinton, ``Imagenet classification with
  deep convolutional neural networks,'' in \emph{Advances in neural information
  processing systems}, 2012, pp. 1097--1105.

\bibitem{everingham2010pascal}
M.~Everingham, L.~Van~Gool, C.~K. Williams, J.~Winn, and A.~Zisserman, ``The
  pascal visual object classes (voc) challenge,'' \emph{International journal
  of computer vision}, vol.~88, no.~2, pp. 303--338, 2010.

\bibitem{ostankovich2018}
V.~Ostankovich and I.~Afanasyev, ``Illegal buildings detection from satellite
  images using googlenet and cadastral map,'' in \emph{International Conference
  on Intelligent Systems}.\hskip 1em plus 0.5em minus 0.4em\relax IEEE, 2018.

\bibitem{girshick2014rich}
R.~Girshick, J.~Donahue, T.~Darrell, and J.~Malik, ``Rich feature hierarchies
  for accurate object detection and semantic segmentation,'' in
  \emph{Proceedings of the IEEE conference on computer vision and pattern
  recognition}, 2014, pp. 580--587.

\bibitem{genitha2010super}
C.~H. Genitha and K.~Vani, ``Super resolution mapping of satellite images using
  hopfield neural networks,'' in \emph{Recent Advances in Space Technology
  Services and Climate Change (RSTSCC), 2010}.\hskip 1em plus 0.5em minus
  0.4em\relax IEEE, 2010, pp. 114--118.

\bibitem{qifang2017super}
X.~Qifang, Y.~Guoqing, and L.~Pin, ``Super-resolution reconstruction of
  satellite video images based on interpolation method,'' \emph{Procedia
  Computer Science}, vol. 107, pp. 454--459, 2017.

\bibitem{neiva2016binary}
M.~B. Neiva, A.~Manzanera, and O.~M. Bruno, ``Binary distance transform to
  improve feature extraction,'' \emph{arXiv preprint arXiv:1612.06443}, 2016.

\bibitem{ronneberger2015u}
O.~Ronneberger, P.~Fischer, and T.~Brox, ``U-net: Convolutional networks for
  biomedical image segmentation,'' in \emph{International Conference on Medical
  Image Computing and Computer-Assisted Intervention}.\hskip 1em plus 0.5em
  minus 0.4em\relax Springer, 2015, pp. 234--241.

\bibitem{kingma2014adam}
D.~Kingma and J.~Ba, ``Adam: A method for stochastic optimization,''
  \emph{arXiv preprint arXiv:1412.6980}, 2014.

\bibitem{DGWorldView3DS2014}
\BIBentryALTinterwordspacing
(2017) Digital globe data sheet for worldview-3 satellite. [Online]. Available:
  \url{https://www.spaceimagingme.com/downloads/sensors/datasheets/DG_WorldView3_DS_2014.pdf}
\BIBentrySTDinterwordspacing

\bibitem{ISBIChallenge}
``Isbi challenge,'' \url{http://brainiac2.mit.edu/isbi_challenge/home},
  (Accessed on 10/29/2017).

\bibitem{long2015fully}
J.~Long, E.~Shelhamer, and T.~Darrell, ``Fully convolutional networks for
  semantic segmentation,'' in \emph{Proceedings of the IEEE Conference on
  Computer Vision and Pattern Recognition}, 2015, pp. 3431--3440.

\bibitem{ioffe2015batch}
S.~Ioffe and C.~Szegedy, ``Batch normalization: Accelerating deep network
  training by reducing internal covariate shift,'' in \emph{International
  Conference on Machine Learning}, 2015, pp. 448--456.

\bibitem{audebert2017joint}
N.~Audebert, B.~L. Saux, and S.~Lef{\`e}vre, ``Joint learning from earth
  observation and openstreetmap data to get faster better semantic maps,''
  \emph{arXiv preprint arXiv:1705.06057}, 2017.

\bibitem{choi2010survey}
S.-S. Choi, S.-H. Cha, and C.~C. Tappert, ``A survey of binary similarity and
  distance measures,'' \emph{Journal of Systemics, Cybernetics and
  Informatics}, vol.~8, no.~1, pp. 43--48, 2010.

\bibitem{russakovsky2015imagenet}
O.~Russakovsky, J.~Deng, H.~Su, J.~Krause, S.~Satheesh, S.~Ma, Z.~Huang,
  A.~Karpathy, A.~Khosla, M.~Bernstein \emph{et~al.}, ``Imagenet large scale
  visual recognition challenge,'' \emph{International Journal of Computer
  Vision}, vol. 115, no.~3, pp. 211--252, 2015.

\end{thebibliography}
%

%








\end{document}